# Fuzzy Logic – Based Scheduling System for Part-Time Workforce

Tri Nguyen[1(✉)] and Kelly Cohen[1]

[1] University of Cincinnati, Cincinnati OH 45221, USA
nguye3hr@mail.uc.edu, cohenky@ucmail.uc.edu

**Abstract.** This paper explores the application of genetic fuzzy systems to efficiently generate schedules for a team of part-time student workers at a university. Given the preferred number of working hours and availability of employees, our model generates feasible solutions considering various factors, such as maximum weekly hours, required number of workers on duty, and the preferred number of working hours. The algorithm is trained and tested with availability data collected from students at the University of Cincinnati. The results demonstrate the algorithm's efficiency in producing schedules that meet operational criteria and its robustness in understaffed conditions.

## 1    Introduction

In 2022, part-time employment constituted nearly 20% of the overall workforce in the United States [1]. Individuals choose part-time positions for various reasons, including flexible scheduling, health considerations, or supplementing income through multiple jobs [2,3,4]. Part-time employees play a crucial role across industries, offering businesses increased productivity and flexibility in workforce allocation. Their presence enhances operational efficiency by allowing employers to adjust work hours based on demand [5]. However, while part-time employment provides numerous advantages, it also presents a significant challenge for employers: scheduling. Employers benefit from the flexibility of creating schedules but must align them with workers' availability and operational requirements. Additionally, scheduling systems must quickly adapt to challenges such as under/overstaffing, unexpected withdrawals, and fairness in shift distribution.

The job scheduling problem has been extensively studied across various sectors, each with unique scheduling criteria. Numerous methods have been proposed to address scheduling challenges, with the most popular approaches being metaheuristic search methods—such as simulated annealing, genetic algorithms, and tabu search—and mathematical programming methods, including linear programming and integer programming. Integer programming has been widely used to optimize schedules for industries with flexible demand and mixed full-time and part-time workforces, such as sales and manufacturing [7,10]. In airline crew scheduling, integer linear programming combined with heuristic search has shown promising results in optimizing shift assignments [13]. In the medical field, the nurse scheduling problem involves creating daily shift schedules while imposing constraints to prevent conflicts and ensure fairness in working hours. Saraswati et al. proposed a modified genetic algorithm to optimize nurse scheduling [9]. Similarly, mixed-integer programming has been applied in healthcare organizations to accommodate flexible demand and employee availability [7]. Other scheduling factors, such as seniority, preferred working hours, and required shift coverage, have been addressed through simulated annealing and variable neighborhood search [8].

Fuzzy logic (FL) is well known for its explainability and ability to incorporate expert knowledge through rule-based systems. Researchers have explored FL for scheduling problems, particularly where uncertainty and vagueness are involved [6,11]. Hai-ting and Lin proposed a task scheduling algorithm for phased array radar systems that uses fuzzy logic to assign priorities. FL has also been widely applied in scheduling problems with unpredictable constraints [16]. In particular, it is often integrated with genetic algorithms (GFS) to learn rules and membership functions, making it a powerful approach for optimizing scheduling tasks. This combination has been successfully applied in workflow scheduling for edge-cloud environments [15] and computational grids [12].

While mathematical programming approaches can yield optimal low-cost solutions, they require significant formulation time and are best suited for problems with well-defined constraints [14]. In contrast, metaheuristic methods, though more adaptable, are computationally expensive, particularly for large search spaces. Unlike these approaches, fuzzy logic is computationally efficient, often providing real-time results. It is also highly flexible, allowing constraints to be embedded within the cost function of genetic algorithms during learning. Given these advantages, this study proposes a fuzzy logic-based model to solve the scheduling problem for a workforce composed entirely of part-time employees. The goal is to generate an optimal schedule that accommodates employee availability, work hour preferences, and the employer's operational requirements.

## 2    The Scheduling Problem

The scheduling scenario involves approximately 20 part-time workers hired each semester for front desk assistant positions within a department. The key constraints and requirements for this scheduling problem are as follows:

- The operation runs five days a week, with an opening duration of 11 hours per day, except for the last day, which operates for 7 hours.

- At least four employees must be on duty at all times to ensure sufficient coverage.

- Each employee submits their weekly availability and preferred working hours per week along with their desired shift length to be considered during scheduling.

- Shift lengths are flexible, but each shift must be at least one hour long to maintain efficiency and avoid excessively short shifts.
- Each employee has a maximum limit on weekly working hours, typically 20 or 25 hours per week, as regulated by institutional policies.
- There is no hierarchy among employees, meaning all workers perform the same tasks, and there are no seniority-based preferences in scheduling.

## 3 Methodology

### 3.1 Fuzzy-Based System to Generate Schedule

Fuzzy logic enables decision-making under uncertainty by modeling human reasoning through linguistic rules. In our approach to solving the scheduling problem, we use a combination of two fuzzy inference systems (FIS1 and FIS2) and an aggregate operator ($\Pi$) to evaluate worker suitability for each shift, as demonstrated in Figure 1.

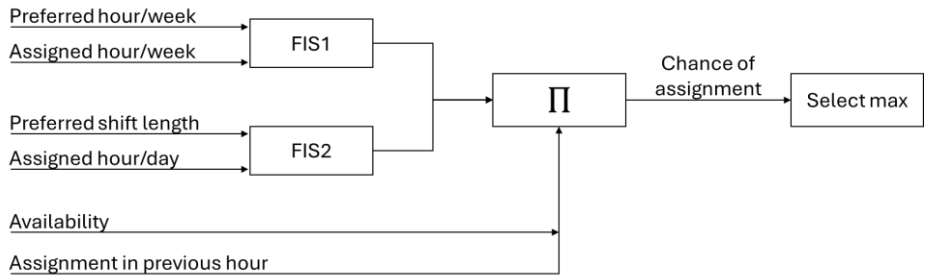

**Fig. 1.** Fuzzy-based shift assignment system.

FIS1 takes an employee's preferred weekly working hours and current total assigned hours as inputs, determining their likelihood of receiving additional shifts. FIS2 evaluates shift length preferences and current assigned hours per day, generating another suitability score. Each FIS produces an intermediate probability of assignment, mimicking human decision-making: initially assigning shifts randomly, then adjusting based on whether assigned hours exceed requested hours. The FISs are Mamdani fuzzy systems that use the AND operator and the centroid method for defuzzification.

The $\Pi$ operator, which performs multiplication, aggregates three key factors: availability, minimum intermediate likelihoods from FIS1 and FIS2, and the previous hour's assignment status. The availability variable is binary—if an employee is unavailable for a shift, its value is zero, resulting in a final probability of zero. The third input, acting as a scaling factor, helps minimize gaps between shifts by prioritizing employees already scheduled in adjacent hours. However, this criterion is not explicitly included in the training process.

Once the final probability of assignment is computed for every employee, they are ranked based on the likelihood, and the top four candidates are selected for the next shift. This process is executed sequentially from the first to the last shift of the week, ultimately generating a complete weekly schedule. The inputs of each FIS share the same predetermined membership functions, which are defined based on expert experience, as shown in Figure 2. In contrast, the output membership functions and rules are learned through genetic algorithms. To reduce training time, FIS1 and FIS2 share the same set of five output membership functions.

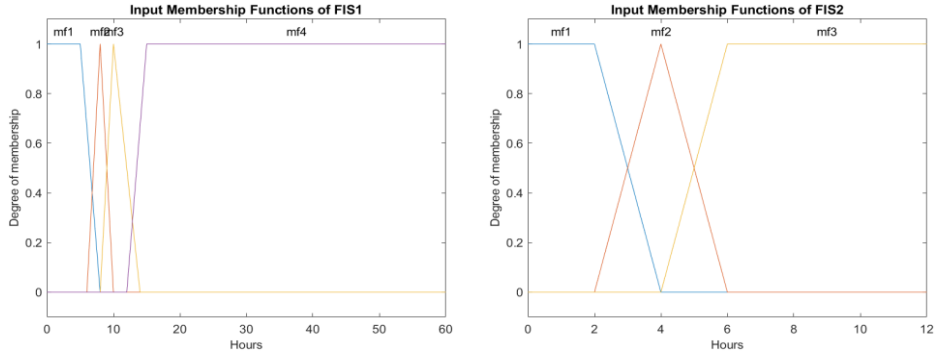

**Fig. 2.** Input membership functions of fuzzy inference systems.

### 3.2 Genetic Algorithm

Genetic algorithms (GAs) are a class of optimization techniques inspired by the principles of natural selection and evolution. The key components of a genetic algorithm include the fitness function and evolutionary operators: selection, crossover, and mutation. GAs enable simultaneous search for solutions in both integer and decimal space. In addition, fitness functions are straightforward to define in this application, making GAs well-suited for learning the FISs.

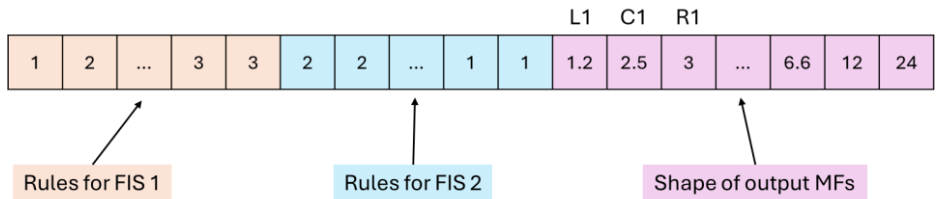

**Fig. 3.** Chromosome structure.

The chromosomes are represented in Figure 2. The first part of the chromosome encodes all possible rules for FIS1 and FIS2, which are used to determine scheduling decisions. The second part encodes the base corners and centers of triangle output membership functions (MFs). Each chromosome represents an individual and a generation consists of multiple chromosomes.

After the first generation is randomly generated, each individual creates schedules based on $n$ different combinations of workers' availability and preferences. A fitness function then evaluates how well the individual performs in generating these $n$ schedules, as follows:

$$\text{Fitness} = \frac{1}{n}\sum_{i=1}^{n}[\text{RMS}(\Delta_1) + \text{RMS}(\Delta_2) + \text{RMS}(\Delta_3)] \tag{1}$$

For each created schedule, the difference between assigned daily hours and preferred shift length ($\Delta_1$), as well as the difference between assigned and preferred weekly hours ($\Delta_2$), is calculated for each worker. If an employee is not assigned a shift on a given day, $\Delta_1$ is not calculated. The $\Delta_3$ term accounts for cases where assignments exceed the permitted weekly working hours, penalizing violations of work-hour limits. The root mean squares (RMS) are then computed for these values, and the average sum over $n$ schedules is used as the fitness score. The fitness function allows the GA to optimize the parameters of the FISs, ensuring that the generated schedules align with employees' preferences.

Following this evaluation, selection is performed to choose parents for the next generation based on their fitness. Common methods include roulette wheel selection, which assigns selection probabilities based on fitness, and tournament selection, where the best among competing candidates is chosen. After parent selection, crossover and mutation are applied to generate the next generation. During crossover, two parents exchange their genes to produce offspring, while mutation introduces small changes in the genes to maintain diversity. This evaluation, selection, and reproduction process continues iteratively until the stopping condition is met.

## 4    Training Setup

Availability data are collected from 70 students at the University of Cincinnati, which is split into 40 students for training and 30 for testing. To train the model, 30 different combinations of workers' availability and preferences are generated. In each combination, 20 workers are randomly selected from the 40 available in the training set. Worker preferences are sampled from a uniform distribution, with weekly work hour preferences ranging between 5 and 15 hours, and preferred shift lengths between 3 and 8 hours. The availability data is represented as a binary 2D matrix (Fig. 4), where each entry indicates whether a worker is available at a given time slot. Preferences for shift length and total weekly work hours are stored in a 1D array of size 20, equivalent to the number of workers. Additionally, a weekly work hour limit of 25 hours is imposed for every worker to ensure compliance with scheduling constraints.

| Day | Hour | Worker Availability (1 = Available, 0 = Unavailable) | | | | | | | | | | | | | | | | | | | |
|---|---|---|---|---|---|---|---|---|---|---|---|---|---|---|---|---|---|---|---|---|---|
| | | A | B | C | D | E | F | G | H | I | J | K | L | M | N | O | P | Q | R | S | T |
| Mon | 8:30 AM - 9:30 AM | 1 | 1 | 1 | 1 | 1 | 1 | 1 | 1 | 1 | 0 | 1 | 1 | 0 | 0 | 1 | 1 | 1 | 1 | 1 | 0 |
| | 9:30 AM - 10:30 AM | 1 | 0 | 1 | 0 | 1 | 1 | 1 | 0 | 1 | 0 | 1 | 1 | 0 | 0 | 0 | 1 | 1 | 1 | 1 | 0 |
| | 10:30 AM - 11:30 AM | 0 | 1 | 0 | 1 | 1 | 1 | 1 | 1 | 1 | 0 | 1 | 1 | 0 | 0 | 1 | 0 | 0 | 0 | 1 | 1 | 0 |
| | - | | | | | | | | | | | - | | | | | | | | | |
| | | Tue, Wed, Thu | | | | | | | | | | | | | | | | | | | |
| Fri | 8:30 AM - 9:30 AM | 1 | 1 | 1 | 1 | 1 | 1 | 0 | 1 | 1 | 1 | 1 | 1 | 0 | 0 | 1 | 0 | 1 | 1 | 1 | 0 |
| | 9:30 AM - 10:30 AM | 0 | 1 | 0 | 1 | 1 | 1 | 1 | 1 | 0 | 1 | 1 | 1 | 0 | 0 | 1 | 0 | 1 | 0 | 1 | 0 |
| | 10:30 AM - 11:30 AM | 0 | 0 | 1 | 1 | 1 | 0 | 1 | 1 | 1 | 1 | 1 | 1 | 1 | 0 | 1 | 0 | 1 | 0 | 1 | 0 |
| | - | | | | | | | | | | | - | | | | | | | | | |

**Fig. 4.** Availability input of workers.

The MATLAB '*ga*' function is configured with specific parameters to optimize scheduling. The population size is set to 200, meaning each generation consists of 200 potential solutions. The algorithm terminates if no improvement is observed for 10 consecutive generations or after 50 generations. Each solution is encoded as a chromosome of size 34, representing rules and output MFs of FIS1 and FIS2. The selection type used is the roulette wheel, which probabilistically selects individuals based on fitness. The EliteCount is set to 10, ensuring the top 10 solutions are directly passed to the next generation without modification. The crossover fraction of 0.8 indicates that 80% of the population undergoes crossover operations and the rest of 20% undergoes mutation.

**Table 1.** Genetic algorithm parameters.

| Parameters | Value |
|---|---|
| Population size | 200 |
| Stall generations | 10 |
| Max generations | 50 |
| Chromosome size | 34 |
| Selection type | Roulette wheel |
| EliteCount | 10 |
| Crossover fraction | 0.8 |
| $n$ | 30 |

# 5    Training Results

| | | Assigned hours per week | | | |
|---|---|---|---|---|---|
| | | Very Low | Low | High | Very High |
| **Preferred hours per week** | Very Low | Low | Very Low | Very Low | Very Low |
| | Low | High | Very Low | Very Low | Very Low |
| | High | Very High | Low | Very Low | Very Low |
| | Very High | Very High | High | Medium | Very Low |

| | | Assigned hours per day | | |
|---|---|---|---|---|
| | | Low | Medium | High |
| **Preferred shift length** | Low | Very High | Very Low | Medium |
| | Medium | Very High | High | Very High |
| | High | Very High | High | Medium |

**Fig. 5.** Rule tables for FIS1 (top) and FIS2 (bottom).

The five membership functions, representing five levels of assignment likelihood (Very Low, Low, Medium, High, and Very High), along with the rule tables for FIS1 and FIS2, are learned through the genetic algorithm and are presented in Figure 5 and Figure 6. The rules for FIS1 learned by the genetic algorithm are intuitive. The likelihood of assignment decreases as more hours are assigned. When the same total number of hours is assigned, the system prioritizes workers who request more hours. The rules for FIS2 are less consistent. While the system tends to prioritize and assign shifts to workers who prefer longer shifts, its behavior is inconsistent when assigned hours per day are high. This discrepancy arises because the rules for lower assigned hours are triggered more frequently during learning, leading to better optimization in lower-hour ranges. Since shift assignments occur in a bottom-up manner, rules governing lower assigned hours are reinforced more effectively. Additionally, because there is no immediate reward for assignments, later assignments may be negatively affected by suboptimal previous assignments (e.g., workers may end up with more hours due to earlier misallocations). Over time, the rules adapt to compensate for these inefficiencies, optimizing shift distribution to mitigate bad initial placements.

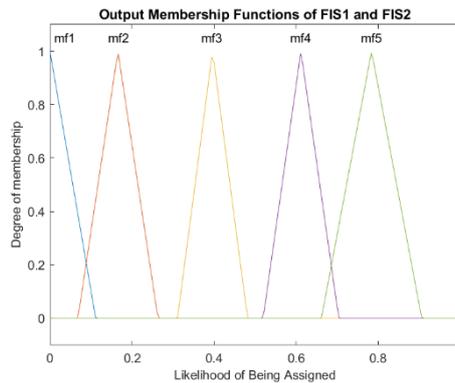

**Fig. 6.** Membership functions of FIS1 and FIS2.

## 6       Testing Results

To evaluate the system, 200 different combinations of availability and preferences are generated from the student testing set. Additionally, although the system is trained by creating schedules for 20 students, another 200 combinations are generated, each including only 16 students, to test the system's robustness. The cost distribution of these testing sets is shown in Figure 7.

The median schedule cost is lower for 20 workers than for 16 workers, indicating that the system produces more optimized schedules when the workforce size aligns with the training conditions. The cost variance is also higher for 16 workers, meaning that the system's performance is less stable under understaffed conditions. Greater difficulty in balancing availability and preferences with fewer personnel is expected.

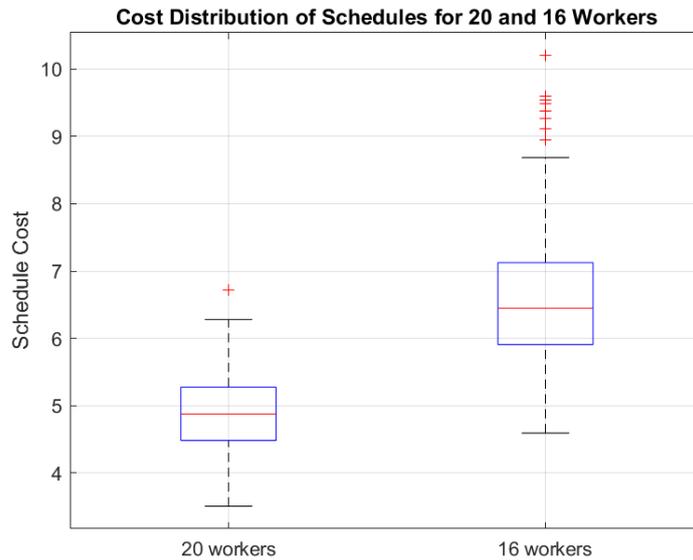

**Fig. 7.** Boxplot of schedule cost distribution from testing results for 20 and 16 workers.

The two schedules at the median cost from the testing results are presented in Figure 8. The key observations are as follows:

- Weekly assigned hours generally align with the requested hours, with only a few workers receiving significantly more or fewer hours than requested.
    - o  In the 20-worker case, two workers are not assigned any hours, while four workers receive three or more additional hours beyond their requests. This tradeoff is reasonable, as the two unassigned workers requested only five hours per week and had low availability.
    - o  In the 16-worker test case, extra weekly hours arise because the total required hours (51 hours × 4 workers) exceed the total requested hours (169 hours), representing an understaffed condition. However, the system successfully allocates shifts to meet operational requirements.

- Extra weekly hours are not evenly distributed among workers. Instead, only a few workers absorb the additional hours. Since the schedule cost remains the same in both cases, the system tends to assign shifts consecutively to the same workers due to the scaling factor for minimizing shift gaps.
- Daily assigned hours exceed preferred shift lengths in some cases, particularly in the 16-worker scenario, where more shifts have three or more additional hours than requested. This discrepancy may result from:
    o The system adapts to the understaffed condition, requiring certain workers to take on longer shifts.
    o Since MFs are predetermined, the input space may not be granular or optimized enough, leading to misalignment between requested and assigned shift lengths.
    o A high scaling factor causes the system to assign consecutive shifts to the same workers, increasing assigned hours beyond preferred shift lengths.
- All assignments remain within the 25-hour weekly limit, ensuring compliance with operational constraints.

No worker is scheduled for more than five days a week, and very few workers are assigned multiple separate shifts in a single day. This confirms that the scaling factor is high, leading the system to prioritize shift continuity over strict adherence to preferred shift lengths.

| Worker ID | | 1 | 2 | 3 | 4 | 5 | 6 | 7 | 8 | 9 | 10 | 11 | 12 | 13 | 14 | 15 | 16 | 17 | 18 | 19 | 20 |
|---|---|---|---|---|---|---|---|---|---|---|---|---|---|---|---|---|---|---|---|---|---|
| Availability | | 90.4% | 73.1% | 73.1% | 78.8% | 76.9% | 71.2% | 61.5% | 73.1% | 67.3% | 67.3% | 78.8% | 59.6% | 65.4% | 73.1% | 76.9% | 80.8% | 65.4% | 71.2% | 84.6% | 63.5% |
| Hour per week | *Requested* | 5 | 12 | 13 | 11 | 15 | 9 | 9 | 10 | 12 | 13 | 14 | 8 | 10 | 12 | 9 | 8 | 8 | 12 | 5 | 5 |
| | Assigned | 6 | 15 | 12 | 11 | 15 | 10 | 13 | 13 | 12 | 16 | 14 | 4 | 13 | 10 | 11 | 9 | 8 | 12 | 0 | 0 |
| | Difference | 1 | 3 | -1 | 0 | 0 | 1 | 4 | 3 | 0 | 3 | 0 | -4 | 3 | -2 | 2 | 1 | 0 | 0 | -5 | -5 |
| | Absolute Difference | 1 | 3 | 1 | 0 | 0 | 1 | 4 | 3 | 0 | 3 | 0 | 4 | 3 | 2 | 2 | 1 | 0 | 0 | 5 | 5 |
| Hour per day | *Requested* | 6 | 7 | 8 | 5 | 6 | 6 | 7 | 3 | 8 | 4 | 3 | 7 | 7 | 3 | 8 | 4 | 6 | 7 | 7 | 8 |
| | Hours assigned on day 1 | 0 | 3 | 0 | 11 | 10 | 0 | 0 | 9 | 4 | 0 | 0 | 0 | 1 | 0 | 0 | 0 | 6 | 0 | 0 | 0 |
| | Hours assigned on day 2 | 0 | 4 | 2 | 0 | 0 | 5 | 5 | 0 | 4 | 7 | 5 | 0 | 3 | 3 | 0 | 0 | 0 | 6 | 0 | 0 |
| | Hours assigned on day 3 | 0 | 0 | 5 | 0 | 0 | 2 | 8 | 0 | 9 | 3 | 0 | 9 | 2 | 0 | 6 | 0 | 0 | 0 | 0 | 0 |
| | Hours assigned on day 4 | 0 | 8 | 5 | 0 | 5 | 0 | 0 | 0 | 1 | 0 | 6 | 2 | 0 | 3 | 5 | 6 | 3 | 0 | 0 | 0 |
| | Hours assigned on day 5 | 6 | 0 | 0 | 0 | 0 | 3 | 0 | 4 | 3 | 0 | 0 | 0 | 2 | 0 | 2 | 0 | 3 | 5 | 0 | 0 |

| Student ID | | 1 | 2 | 3 | 4 | 5 | 6 | 7 | 8 | 9 | 10 | 11 | 12 | 13 | 14 | 15 | 16 |
|---|---|---|---|---|---|---|---|---|---|---|---|---|---|---|---|---|---|
| Availability | | 71.2% | 84.6% | 76.9% | 59.6% | 73.1% | 78.8% | 73.1% | 90.4% | 65.4% | 61.5% | 88.5% | 80.8% | 73.1% | 78.8% | 65.4% | 76.9% |
| Hour per week | *Requested* | 8 | 8 | 12 | 5 | 15 | 13 | 12 | 11 | 11 | 8 | 14 | 10 | 13 | 13 | 7 | 9 |
| | Assigned | 14 | 16 | 10 | 6 | 19 | 14 | 12 | 11 | 18 | 6 | 15 | 13 | 16 | 19 | 5 | 10 |
| | Absolute Difference | 6 | 8 | 2 | 1 | 4 | 1 | 0 | 0 | 7 | 2 | 1 | 3 | 3 | 6 | 2 | 1 |
| Hour per day | *Requested* | 6 | 7 | 8 | 5 | 6 | 6 | 7 | 3 | 8 | 4 | 3 | 7 | 7 | 3 | 8 | 4 |
| | Hours assigned on day 1 | 0 | 0 | 10 | 0 | 11 | 0 | 6 | 11 | 1 | 0 | 0 | 5 | 0 | 0 | 0 | 0 |
| | Hours assigned on day 2 | 7 | 5 | 0 | 0 | 0 | 4 | 6 | 0 | 4 | 0 | 4 | 3 | 2 | 2 | 5 | 2 |
| | Hours assigned on day 3 | 0 | 0 | 0 | 0 | 10 | 0 | 0 | 0 | 2 | 1 | 11 | 0 | 3 | 10 | 0 | 7 |
| | Hours assigned on day 4 | 6 | 11 | 0 | 0 | 1 | 0 | 0 | 0 | 11 | 5 | 0 | 5 | 5 | 0 | 0 | 0 |
| | Hours assigned on day 5 | 1 | 0 | 0 | 6 | 7 | 0 | 0 | 0 | 0 | 0 | 0 | 0 | 6 | 7 | 0 | 0 |

**Fig. 8.** Worker schedules corresponding to the median schedule cost for 20 (top) and 16 (bottom) workers.

## 7 Conclusion

This study presents a fuzzy logic-based scheduling system optimized with genetic algorithms to efficiently allocate shifts to part-time workers while balancing availability, work preferences, and operational constraints. The system utilizes two Fuzzy Inference Systems (FIS1 and FIS2) and a multiplicative operator to determine assignment likelihoods. The GA optimizes fuzzy rule tables and output membership functions to enhance scheduling efficiency. Testing on 200 scenarios demonstrates that ideal workforce sizes (20 workers) result in lower schedule costs and more balanced shift distributions, whereas fewer workers (16 workers) lead to increased variance and shift over-assignments due to understaffing. However, the system shows high adaptation and ensures compliance with the operational requirements.

Future improvements include training with variable workforce sizes, tuning input membership functions and scaling factors, and refining the fitness function to better handle fairness in shift assignments. Additionally, incorporating a dedicated fitness function for shift length and daily hours could allow for more flexible schedules with break periods. Given the high turnover rate (approximately 40%), new features - such as scheduling new hires to work the same shift as senior employees for better shadowing and support - can be implemented. Finally, integrating a hybrid approach where the deterministic schedule is further optimized by genetic algorithms could improve overall scheduling efficiency.